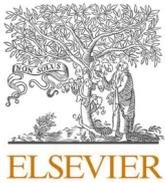

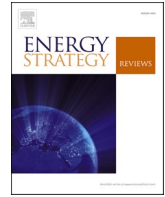

# Novel application of Relief Algorithm in cascaded artificial neural network to predict wind speed for wind power resource assessment in India

Hasmat Malik [a], Amit Kumar Yadav [b,*], Fausto Pedro García Márquez [c], Jesús María Pinar-Pérez [d]

[a] Berkeley Education Alliance for Research in Singapore (BEARS) (a research centerof the University of California, Berkeley, USA), Singapore
[b] Electrical and Electronics Engineering Department, National Institute of Technology, Sikkim, Ravangla, Barfung Block, South Sikkim, 737139, India
[c] Ingenium Research Group, Universidad Castilla-La Mancha, 13071, Ciudad Real, Spain
[d] Department of Quantitative Methods, CUNEF Universidad, 28040, Madrid, Spain

## A B S T R A C T

Wind power generated by wind has non-schedule nature due to stochastic nature of meteorological variable. Hence energy business and control of wind power generation requires prediction of wind speed (WS) from few seconds to different time steps in advance. To deal with prediction shortcomings, various WS prediction methods have been used. Predictive data mining offers variety of methods for WS predictions where artificial neural network (ANN) is one of the reliable and accurate methods. It is observed from the result of this study that ANN gives better accuracy in comparison conventional model. The accuracy of WS prediction models is found to be dependent on input parameters and architecture type algorithms utilized. So the selection of most relevant input parameters is important research area in WS predicton field. The objective of the paper is twofold: first extensive review of ANN for wind power and WS prediction is carried out. Discussion and analysis of feature selection using Relief Algorithm (RA) in WS prediction are considered for different Indian sites. RA identify atmospheric pressure, solar radiation and relative humidity are relevant input variables. Based on relevant input variables Cascade ANN model is developed and prediction accuracy is evaluated. It is found that root mean square error (RMSE) for comparison between predicted and measured WS for training and testing wind speed are found to be 1.44 m/s and 1.49 m/s respectively. The developed cascade ANN model can be used to predict wind speed for sites where there are not WS measuring instruments are installed in India.

## 1. Introduction

Inadequate sources of energy have become one of the major barriers realized by the global sustainable development. Therefore renewable energy sources have drawn attention worldwide. Among renewable energy sources, wind energy (WE) will act as important source for future electricity in the form of wind power. The demand of wind energy worldwide is developing drastically [1] so accurate assessment of WE potential [2–5] is important for economic point of view. Since wind power is a function of wind speed (WS) [6], prediction of power are generally derived from prediction of speed as shown in Eq. (1).

$$P = \frac{1}{2}\rho A v^3 \tag{1}$$

where $A$ = Rotor swept area, $\rho$ = air density, $v$ = velocity of wind in m/s.

The greatest problem with wind power integration in power system is fluctuating nature of wind power due to high correlation with stochastic nonstationary WS. This escalate the essential for reliable and accurate techniques for wind farms output power prediction. WS prediction is also essential process for wind farms units' maintenance, optimal power flow between conventional units and wind farms [7], electricity marketing bidding, power system generators scheduling, energy reserve, storages planning and scheduling. Several physical model [8,9] and time-series-based models have been investigated for WS forecasting [10–15].

The three main objectives of this review are as follows: (i) to present different Artificial Neural Network used in short, medium and long term wind speed prediction (ii) to summarize important findings in WS prediction field (iii) to find out research gaps in WS prediction field.

The paper is organized as follows. Review and classification of WS

* Corresponding author.
*E-mail addresses:* hasmat.malik@gmail.com (H. Malik), amit1986.529@rediffmail.com (A.K. Yadav), faustopedro.garcia@uclm.es (F.P.G. Márquez), jesusmaria.pinar@cunef.edu (J.M. Pinar-Pérez).

forecasting methods is given in section 2. It includes different approaches utilized for very short, medium and long term WS prediction along with most relevance parameters selection. Important findings are presented in section 3. Section 4 presents the research gaps identified in this study. Finally, conclusions are given in section 5.

## 2. Review and classification of wind speed prediction methods

There are numerous methods and techniques that have been developed for prediction of WS. To attain stable power in interconnected power system operations, it is necessary to predict WS. Physical, statistical and combination of same or different approaches are the present WS prediction methods which is also classified on the basis of various time scales. Short term forecasts include forecasting WS for several days and also from minutes to hours [16]. The long term forecasts means longer time scale prediction up to a month, a year or several years [17].

### 2.1. Short and extremely short term wind speed prediction

Extremely short and short term WS forecasting models predict WS of small duration from few seconds to 6 h. Statistical, time series and physical approach are used for short forecasting [18–24]. It is mainly helpful for unit commitment and it also reduces the voltage and frequency fluctuations due to variation in wind power.

Li et al. [25] employed Elman recursion neural network to predict one step advance WS. Ten minutes previous data is used to forecast one step ahead prediction. The network is tested with different number of neurons. It is found that error prediction with Weibull distribution is low in comparison to Gaussian distribution. It is found that Normalized mean square error (NMSE) using Elman network is 0.0166 and with BP network is 0.0169.

Cao et al. [26] presented recurrent neural networks using the measured data at different heights to predict WS. The results obtained with multivariate recurrent neural network are compared to the univariate recurrent neural network and ARIMA model. At different heights univariate ARIMA, univariate ANN, multivariate ARIMA, multivariate ANN are discussed. It is concluded that recurrent neural network (RNN) model preferred over ARIMA for univariate and multivariate case.

Da Liu et al. [27] found that accurate and effective results are obtained with this approach. WS data and temperature are taken from a wind farm of china are used for training and testing the model. Model output is compared with ARIMA, generalized regression neural network and echo state networks (GRNN) and echo state networks (ESN) model. Prediction at an interval of 15 min (short term) is made by spectral clustering and echo state networks (SC-ESN). The performance of the model is also tested with data collected from CEDA, UK. Mean values of MAPE%, MAE and normalized root mean square error (NRMSE) are found to be 11.76, 0.40 and 1.38 resp. with SC-ECN-GA model.

Kusiak and Zhang [28] presented double exponential time series (DES) and NN, support vector machine (SVM), k-nearest neighbour, boosting tree random forest data driven models to forecast short term WS and power. WS for six time intervals from t+10 s to t+60 s are predicted and found that DES and NN model outperforms the other four models. For WP prediction three models are established and each model is trained with five algorithms (NN, boosting tree, random forest, SVM, and k-nearest neighbour) to predict short time horizon power.

Short term WS is predicted using SVM model based on granulated information that obtained with fuzzy information granulation (FIG) by Xiao Cheng and Peng Guo [29]. The redundant information from dataset reduces with fuzzy information granulation and hence predicts WS with a good accuracy. The data series measured from wind farm in Zhangjiakou is used to evaluate performance of the model. It is found that using combined SVM and FIG approach predicted range of variation of Low, R and Up (fuzzy granule parameters) are 7.8838, 8.7280 and 10.2073 respectively.

Khalid and Savkin [30] use data from turbines and numerical weather predictor (NWP) to improve the model accuracy. For testing model, two data sets of different time horizons with different turbine groups have been used. Predicted WS and direction are used to predict output power with improvement from existing models. MAE and root mean square error (RMSE) values obtained with 1st data set are 2.27% and 2.69% and with 2nd data set are 2.87% and 3.27%.

Cadenas and Rivera [31] developed ANN model to forecast WS at different time intervals. The MAPE varies from 0.0399 to 0.0449. To forecast WS different time intervals are taken as shown bellow

$$WS_{t+1} = f(WS_t, WS_{t-1}) \qquad (2)$$

$$WS_{t+2} = f(WS_{t+1}, WS_t) \qquad (3)$$

$$WS_{t+3} = f(WS_{t+2}, WS_{t+1}) \qquad (4)$$

Sheela and Deepa [32] proposed a hybrid model by combining Self Organizing Maps (SOM) and multilayer perceptron (MLP) neural network to predict wind speed more accurately. WS, wind direction and temperature are the input. Similar groups of all the input data are formed by SOM and then MLP network is used to predict WS. It is found that performance of hybrid model is better with RMSE 0.0828 as compared to conventional MLP, back propagation network (BPN) and radial basis function network (RBFN) models. RMSE obtained with proposed hybrid approach is 0.0828.

Two hybrid methods ARIMA-ANN and ARIMA-Kalman are employed to forecast WS and their performance is compared by Liu et al. [33]. A time series ARIMA is used with ANN model to predict the best structure for network and also to find the best initial conditions with Kalman model. The model is used for multistep ahead WS. Broyden-Fletcher–Goldfarb–Shannon Quasi-Newton Back Propagation (BFGS) is the training algorithm use with ANN model. It is found that WS predicted by ARIMA-Kalman model is better than ARIMA –ANN model. The improved in MAE from 1 to 3 step is 53.47%, 47.40% and 49.18% resp. Improvement in MAPE error is found to be 52.74%, 46.61% and 48.41% respectively.

El-Fouly et al. [34] presented Grey predictor model GM (1, 1) to predict WS and WP. Based on the forecasted WS as input to a 600-W wind turbine, output power is predicted. For 1 h ahead prediction the proposed model is tested with hourly recorded datasets. The results obtained with Grey predictor model are found to be better than persistent model with an average accuracy improvement of 11.24% and 12.20% for WS and power output respectively. Niu and Wang [35] proposed empirical mode decomposition method along with multi-objective optimization algorithm to forecast WS for three different sites in China. It is suggested that proposed model give better results than traditional model. Zhou et al. [36] considered data analysis, model selection strategy, forecasting processing combined with a modified multi-objective optimization algorithm, and model evaluation for forecasting WS in Gansu, China. This approach improve WS forecasting accuracy. Jiang et al. [37] removed noise using decomposition technique and proposed hybrid linear-nonlinear modeling with chaos theory. Finally support vector regression and firefly optimization algorithm are combined to predict WS.

Hao et al. [38] proposed hybrid of clustering algorithm and generalized regression neural network (GRNN) for WP prediction. Dilation and erosion, K-mean, DPK- medoids are used to cluster inputs variable having similar characterstics. Normalized root mean square error for no clustering, DE, DPK- medoids and K-means are found to be 13.9%, 7.9%, 13% and 8.9% respectively.

Liu et al. [39] developed hybrid model by combining singular spectrum analysis to decompose WS into number of components, Convolutional Neural Network to predict trend component and Support vector regression for detail component prediction. Computation time for hybrid model is under 3 min. Some other important researches on WS prediction with and without hybrid models are given in Refs. [40–66].

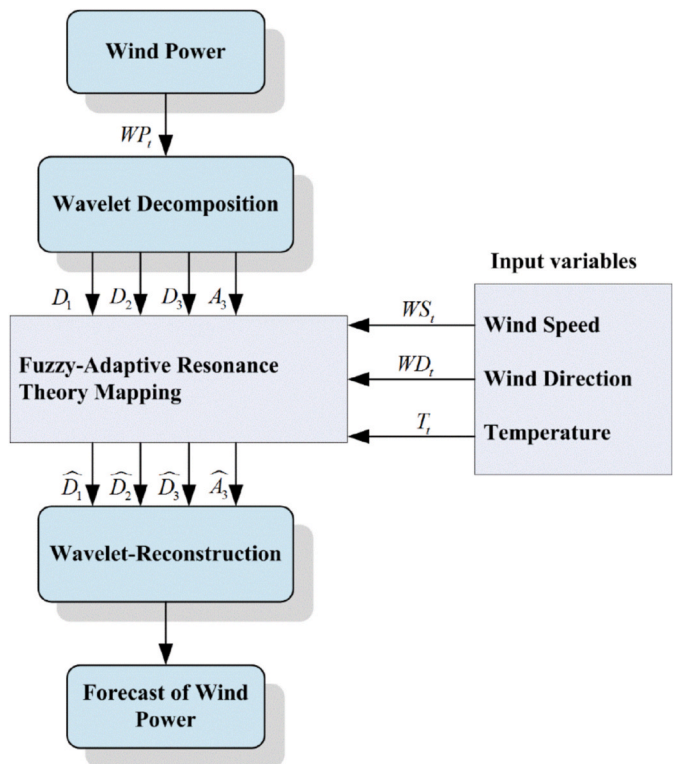

**Fig. 1.** Forecasting wind power fuzzy model.

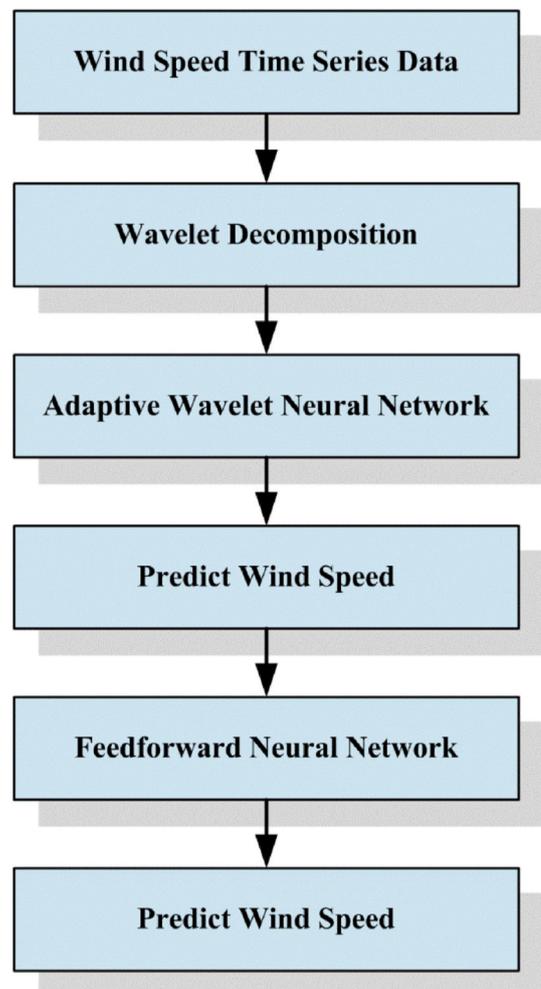

**Fig. 2.** WS prediction using Wavelet Decomposition.

### 2.2. Medium term wind speed prediction

Forecasting models of medium duration include 6 days to 1 day ahead prediction. It is helpful for taking generator online and offline decisions. These models are self-calibrating and self-adjusting. It includes time series, statistical and physical models of WS prediction [67–75].

Maatallah et al. [76] developed a model based on Hammerstein Auto-Regressive (HAR) model and verified the result with real time data of two sites one at Illinois and Other is at New Jursey. The HAR model outperforms the ARMA and ANN models but it is also found that at some data point error was high. HAR model nonlinearity generates these errors. These errors can be overcome by using hybrid framework with HAR model. Result demonstrates that HAR model performs well than ARMA and HAR model.

Haque et al. [77] presented a hybrid machine learning algorithm to forecast WP on the basis of data collected from a wind farm of Canada. The hybrid approach is based on wavelet transform and fuzzy adaptive resonance theory mapping (ARTMAP) [Fig. 1.]. The model is proposed for different k hours ahead WS prediction. The results obtained with machine learning algorithm are found to be better than persistence and numerically weather prediction models. Further improvement can be done by taking more parameters e.g. humidity, pressure and precipitation etc into consideration to predict WS. For daily WP forecasting using proposed approach average MAPE, normalized root mean square error (NRMSE) and normalized mean absolute error (NMAE) are found to be 12.73%, 2.05% and 1.52%.

Hu et al. [78] incorporated hybrid approach Empirical Wavelet Transform- Coupled Simulated Annealing- Least Square SVM Support Vector Machine (EWT-CSA-LSSVM) to forecast accurate short term WS. New wind series is formed after decomposing the real wind series with EWT. This reconstructed wind series is used to predict WS with LSSVM. The Coupled Simulated Annealing (CSA) is the algorithm used to optimize parameters of LSSVM. The approach is proposed to forecast results with different kernel functions. Best result is obtained with linear mapping function. The accuracy is better but this approach takes computation time more than persistence method and AR model. The proposed model is employed for multi-step and one step ahead time horizons WS prediction. Using proposed approach for one step forecasting horizon RMSE, MAE, MAPE indicator values are found to be 0.57, 0.57 and 12.89 respectively. Indicators for multistep forecasting also shows that proposed approach gives promising results in WS forecasting area.

Da Liu et al. [79] proposed Wavelet transforms and support vector machine optimized with genetic algorithm (GA) for prediction. Granger causality test is used to select proper lags of temperature in input variables to SVM. The performance of this model is compared and found to be better than persistent and SVM-GA model. Humidity, air pressure, precipitation and other input variables in addition with temperature make SVM better performance. MAE, MAPE% and RMSE are used to evaluate performance of proposed approach and their values found to be 0.6169, 14.79 and 1.2234.

Dan-dan et al. [80] used SVM particles optimized with particle swarm optimization (PSO) to predict short term WS. It is found that the accuracy to predict WS with combine SVM model is more than single linear and combined traditional linear forecasting approaches. With PSOSVM approach Mean Absolute Deviation (MAD), MSE, Mean Forecast Error (MFE) and MAPE values are found to be 0.6013, 0.7184, 0.0012 and 15.0017 respectively. Some other important researches on medium WS prediction are given in Refs. [81–103].

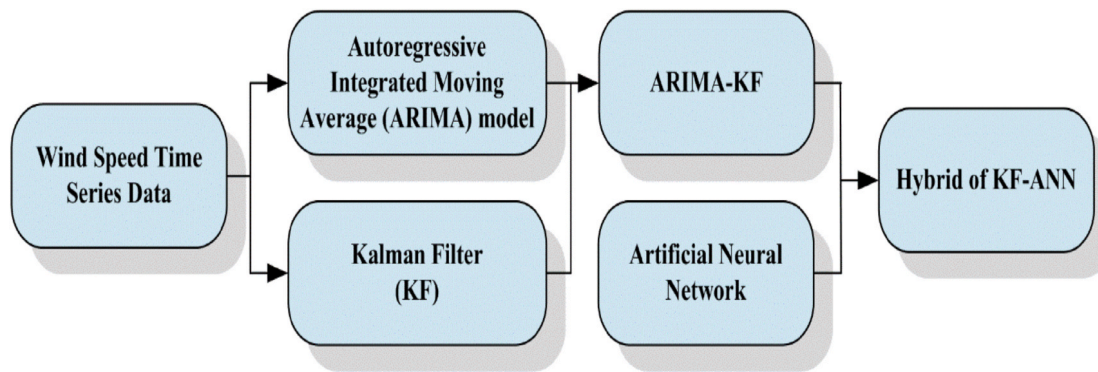

**Fig. 3.** KF-ANN model.

## 2.3. Long term wind speed prediction

Long term forecasting models include 1 day to 1 week or more ahead prediction. Long term WS can be predicted with physical, statistical, time series approaches. It is useful for planning a new wind turbine and for deciding the location of a new wind power plant. Guo et al. [104] used hybrid SARIMA-LSSVM approach for monthly mean WS prediction in Hexi Corridor, China. Error due to the nonlinearity and seasonal variations are reduced by this hybrid approach. Results obtained with proposed hybrid technique are compared with traditional forecasting methods and it is found that proposed approach gives accurate and efficient wind energy prediction results. The MSE, MAE and MAPE% values using proposed approach for Mazong Mountain area found to be 0.12, 0.30, 6.76 and for Jiuquan are 0.11, 0.26, 10.49 respectively. Barbounis and Theocharis [105] employed locally recurrent neural network to forecast WS in China. The online learning algorithm based on recursive prediction error (RPE) is used for training and this algorithm improves the performance of model. IIR-MLP model trained by GRPE shows 62.70% and 70.22% improvement in WS and power measurement. Mohandes et al. [106] developed SVM approach to predict WS. Daily WS data of Madina city, Saudi Arabia is used for this technique. The predicted values are compared with MLP neural networks and it is found that SVM approach outperforms the MLP model with 1–11 orders. The lowest MSE obtained with SVM model with data order 11 is 0.0078. Bhaskar and Singh [107] proposed adaptive wavelet neural network and feed forward neural network to predict WS and wind power. The proposed algorithm is shown in Fig. 2. MAE and RMSE values from 1 h to 30 h ahead WS prediction are found to be 7.081 and 10.221 respectively (see Fig. 3).

Taylor et al. [108] used ensemble predictions and time series models to forecast wind power density. WS data from five wind farms Blood Hill in Nzorfolk, Llyn Alaw in Anglesey, Bears Down in Cornwall, Bu Farm, Orkney, and Cemmaes in Powys is used for study. Wind power density forecasted with weather ensemble predictions outperforms the other time series models such as autoregressive moving average-generalized autoregressive conditional heteroskedasticity (ARMA-GARCH) models, autoregressive fractionally integrated moving average-generalized autoregressive conditional heteroskedasticity (ARFIMA-GARCH) models. Velo et al. [109] presented MLP neural network supervised with back propagation (BP) learning algorithm to predict annual average WS. The data of WS and WD of nearby sites are used as input parameters to the proposed model. The RMSE is 1.31. Guo et al. [110] developed a new technique based on the chaotic time series modelling and Apriori algorithm for prediction. Data from four meteorological stations of China are used for testing and verification of model. The proposed model performance is compared with the ARIMA, ANN and weighted local-region (WLR) methods. It is found that minimum absolute relative error (ARE) is obtained with the proposed model. Its value for four sites ranges from 16.94 to 27.50.

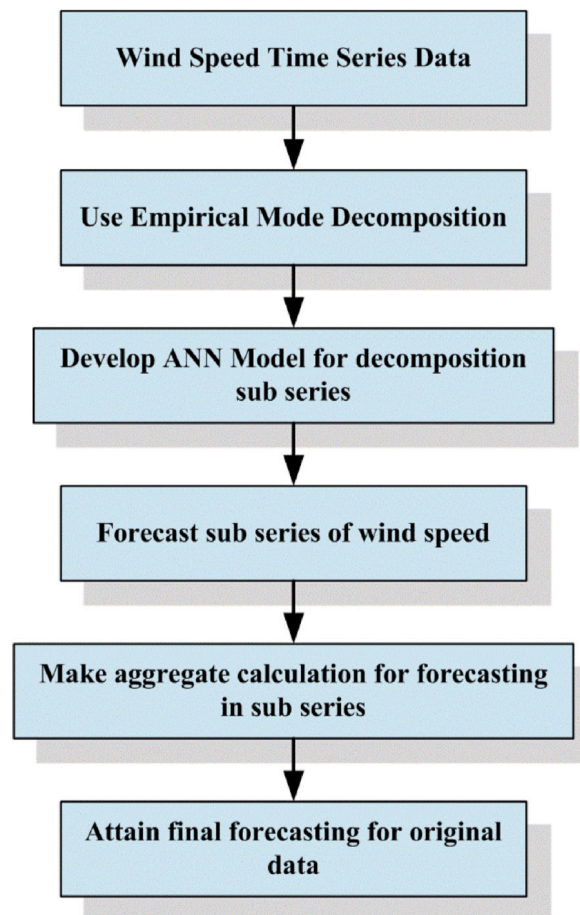

**Fig. 4.** Hybrid EMD-ANN

Shukur and Lee [111] utilized kalman filter (KF) and ANN for improving WS forecasting. ARIMA decide input structure for ANN and KF. Hybrid KF-ANN forecast better WS than its separate components.

Ling-ling et al. [112] predicted WS more effectively by using wavelet theory along with ARMA model. To reduce errors in predicted values low frequency data from collected data series is constructed using wavelet theory and then ARMA model is use to predict wind speed. It is found that as compared to single time series model, performance of the proposed approach is better. Liu et al. [113] modified Taylor Kriging method and applied to predict time series WS. The performance of proposed approach is compared with ARIMA method for WS data collected from a site of USA. It is found that proposed approach is more accurate and improves the output with average value of MAE and RMSE to 18.60% and 15.23%. Arjun et al. [114] used Multivariate Regression models to predict WS using independent variables as air temperature, relative humidity and pressure. Using Linest and Logest functions of Excel four models are designed namely Linest model 1, Linest model 2, Logest model 3 and Logest model 4. The least sum of square errors obtained with model three is 121.4 but the F-statics value for model 4 is highest among all other models. It is also concluded that further improvement and analysis is required to predict WS more accurately. Wang et al. [115] used hybrid empirical mode decomposition (EDM) and Elman neural network (ENN) for accurate WS forecasting. The collected hourly data series from a site of China is divided into sub series using EMD and then ENN model is established using these sub series. Model performance is compared with Persistent Model (PM), BPNN and ENN models and found that accurate and effective wind speed is predicted with hybrid approach. This conclusion is obtained on the basis of calculated MAE, MSE and MAPE values of four seasons. Barbounis et al. [116] developed three recurrent neural networks to forecast WS at 72 time steps ahead. Two novel learning algorithms namely global and decoupled recursive prediction error are introduced for the training of the recurrent forecast models. These learning algorithm has small error.

**Table 1**
Summary of the most important articles applying data decomposition methods in wind energy applications.

| Reference | Data Decomposition Technique | Predictor Techniques |
|---|---|---|
| Liu et al. [131] | EMD | ARIMA, BPNN, ELM, ENN, GRNN |
| Wang et al. [132] | Singular spectrum analysis | Hybrid Laguerre neural network |
| Liu and Duan [133] | wavelet | Ensemble model |
| Ma et al. [134] | ensemble empirical mode decomposition | LSTM |
| Peng et al. [135] | wavelet soft threshold denoising | gated recurrent unit |
| Deng et al. [136] | empirical wavelet transform | Elman neural network |
| Memarzadeh et al. [137] | wavelet transform | LSTM |
| Moreno et al. [138] | variational mode decomposition and singular spectral analysis | long short-term memory neural network, adaptive neuro-fuzzy system, echo state network, support vector regression and Gaussian regression process |
| Mi et al. [139] | Singular Spectrum Analysis | Modified Adaptive Structural Learning of Neural Network |
| Lu et al. [140] | ensemble empirical mode decomposition | grey-box model |
| Abedinia et al. [141] | improved wavelet transform | two-dimensional convolution neural network |
| Abedinia et al. [142] | improved version of empirical mode decomposition | bagging neural network and K-mean clustering |
| Gu et al. [143] | variational mode decomposition | radial basis function neural network |
| Jiajun et al. [144] | wavelet transform | ensemble learning and deep belief network |
| Liu et al. [145] | real-time decomposition | Bidirectional Long Short-Term Memory |
| Peng et al. [146] | optimal variational mode decomposition | regularized extreme learning machine ensemble model |
| Wu et al. [147] | complete ensemble empirical mode decomposition | extreme learning machine |
| Zhang et al. [148] | variational mode decomposition | Hybrid of Principal component analysis- radial basis kernel Function, Auto-regressive and moving average |

Atwa and El-Saaday [117] employed constrained grey predictor model to predict annual WS. The grey predictor model use clustered data instead of time series based data sets. To validate the model, data from a site is used to test the model. The proposed model outperforms the Weibull probability density function (PDF) model with annual percentage average absolute error of 13.4%. Liu et al. [118] developed hybrid of Empirical Mode Decomposition and ANN for WS prediction. It is found that EMD-ANN model outperforms ANN, ARIMA and SM [Fig. 4].

A new model Cuckoo Search-Fuzzy System-Weather research and forecasting ensemble (CS-FS-WRF-E) is proposed by Jing Zhao et al. [119]. It is a new approach and based on NW Pensemble and a fuzzy process. The performance of proposed approach is compared to mean WRF-E (M-WRF-E), fuzzy system WRF-E (FS-WRF-E) and CS–FS–WRF-E. MSE obtained for these approaches are 7.4774, 7.3615 and 3.2847 respectively. Linear models are combined with wavelet decomposition technique for a week ahead wind prediction by Dennis C. Kiplangat et al. [120]. The model performance is evaluated on 10 min data samples of 234 locations collected from National Renewable Energy Laboratory (NREL), USA. Average error using proposed approach for 3 days ahead prediction is within 7–8%. Accuracy is improved up to 18.23% for week ahead wind speed prediction. Some other important researches on long term WS prediction with network types are given in Refs. [121–130]. Furthermore study based on data decomposition with different predictor is shown in Table 1.

### 2.4. Relevant input parameters selection for wind speed prediction

The various geographical and meterlogical variables as input parameters to different models play an important role in WS prediction. To forecast WS accurately it is also required to remove seasonal variations from input data. It is quite important to mention here that effect of all input variables to predictive model is not same. Therefore it is the primary requirement of researcher to choose most relevant parameters to reduce error in forecasted values [149].

Luis Vera-Tudela and Martin Kühn [150] discussed different input selections methods to choose predictors from a set of input variables. Six sets with different number of predictors are formed. First set consists of all 56 input variables and is a baseline set. The 23 predictors are used in second set based on the ranking of spearman coefficient calculated for each input variable. Stepwise regression is used for selection of third predictor set. Redundant candidates are eliminated from the fourth set with cross correlation higher than 0.95. Fifth set is formed using clustering and consist 17 predictors and sixth set consists of principal components. An optimal set of predictors is formed four filters and one dimensional reduction algorithm.

Similarly a hybrid network with Bayesian clustering by dynamics (BCD) and support vector regression (SVR) is employed to forecast 48 h ahead wind power generation by Shu Fan et al. [151]. Data from wind farm installed in United States is used for evaluating the performance of model. Wind generation and WS data from the available input variables are used for SVR model whereas wind direction and humidity along with WS and wind generation are used for training and testing of BCD performance. The selection of network parameters makes the proposed method effective to forecast wind generation.

An IS-PSO-BP method is employed by Chao Ren et al. [152] for WS forecasting. The optimal input parameters are selected for the back propagation neural network based on Particle Swarm Optimization (PSO) for wind speed prediction. Forecasting accuracy is increased with input selection method. The model performance is evaluated with the WS data of Jiuquan prefecture and Yumen city. Lateral and longitudinal dataset selection methods are used and it is found that longitudinal data selection method is more useful and suitable for WS prediction. The proposed model performance is compared with BP and ARIMA model. Wu et al. [153] proposed day ahead forecasting model consist of deep feature extraction, clustering and long short-term memory (LSTM). Deep feature extraction improve forecasting accuracy by 33% in comparison to principal component analysis.

## 3. Important findings

Through comprehensive literature review, it is obtained that different physical, statistical and hybrid approaches are available for WS prediction. ARMA model, Time series models, Spatial correlation, Artificial intelligence etc. are the commonly used techniques. Different types of soft computing models used for prediction and forecasting are Elman recursion neural network, Recurrent neural network, support vector machine, ANN with LM algorithm, Probabilistic fuzzy system, Radial basis function neural network, wavelet support vector machine, Recurrent neural network, Adaptive wavelet neural network and multi resolution, ANFIS, Support vector regression, Recurrent multilayer network with internal feedback paths, Chaotic network operator, adaptive wavelet neural network, etc. The hybrid methods used for WS forecasting are Self Organizing Maps and MLP, ARIMA- ANN and ARIMA-Kalman filter, Piecewise support vector machine (PSVM) and SVM, wavelet transform and fuzzy, Empirical Wavelet Transform- Coupled Simulated Annealing- Least Square Support Vector Machine, Wavelet transforms and support vector machine optimized with genetic algorithm, SVM particles optimized with particle swarm optimization, Wavelet decomposition and artificial bee colony algorithm, Adaptive weighted particle swarm optimization (AWPSO) and ANN, empirical mode decomposition (EDM) and Elman neural network (ENN), ENN and GRNN, EMD-ANN, PSO and support vector machine regression, Bayesian clustering by dynamics (BCD) and support vector regression (SVR).

These techniques have different forecasting accuracy that mainly decided on the basis of MSE, MAPE, MAE (most commonly used error values). Forecasting accuracy further improves with relevant input parameters selection and seasonal component adjustment. It is found that for short term horizons NNs, ARIMA, SVM are generally used techniques whereas for long term prediction hybrid approaches are applied. Although this is not the crucial classification. Hybrid approaches can also be implied for short term WS prediction and statistical approaches for long term WS forecasting. ARIMA is used to forecast WS and the error due to nonlinearity of ARIMA model is overcome by ANN. Prediction accuracy of Reduced support vector machine model increases with wind pressure and temperature as inputs but with wind direction along with other parameters prediction accuracy decreases.

Accuracy of forecasting model mainly depends on input parameters used for the study. It is also found that insertion or deletion of one or more parameters effects model prediction results. The optimal input parameters are selected for the back propagation neural network based on Particle Swarm Optimization (PSO) for WS prediction. For prediction air temperature, vapor pressure, relative humidity, pressure, wind direction and evaporation are commonly used input variables. In some references latitude, longitude, mean sea level pressure, dew point temperature, time of day are also included in input parameters for WS prediction.

## 4. Identified research gaps in wind speed prediction

Based on literature review the following research gaps are identified in the present study:

(1) A more comparative analysis is required to develop a model that does not affected by time horizons of input data. Such that same model can be used to forecast short term as well as long term wind speed effectively and accurately.

(2) Some methods should be developed to calculate the best performance of ANN model with required number of neurons in hidden layer instead of evaluating the model performance by varying these neurons one by one.

(3) More studies are required to find out the effect of seasonal variations on performance of input variables to different techniques.

**Table 2**
Description of 12 locations of different, Indian states.

| S. No. | Locations | State | ANN Model |
|---|---|---|---|
| 1 | Mount Harriet | Andaman and Nicobar | Testing |
| 2 | South Bay | | Training |
| 3 | Burgula | Andhra Pradesh | Training |
| 4 | Chinnakaballi | | Training |
| 5 | Kotrathanda | | Training |
| 6 | Galikonda | | Training |
| 7 | Kotturu | | Training |
| 8 | Shahpuram | | Training |
| 9 | Siddanagatta | | Training |
| 10 | Singarikonda | | Training |
| 11 | Talaricherevu | | Training |
| 12 | Teranavalle | | Training |
| 13 | Tirumalayapalli | | Training |
| 14 | Vajrakarur 2 | | Testing |
| 15 | P.Leikul | Assam | Training |
| 16 | Dhrobana | Gujarat | Training |
| 17 | Khambada | | Training |
| 18 | Lamba | | Training |
| 19 | Mahidad | | Training |
| 20 | Rojmal 2 | | Training |
| 21 | Sadodar | | Training |
| 22 | Sangasar | | Training |
| 23 | Sinugra | | Training |
| 24 | Suvarda | | Training |
| 25 | Vadgam | | Training |
| 26 | Vandhya | | Training |
| 27 | Lodhrani | | Testing |
| 28 | Bidda | Jammu and Kashmir | Training |
| 29 | Hulkoti | Karnataka | Training |
| 30 | Jogimatti | | Training |
| 31 | Mannikeri | | Training |
| 32 | Mavingundi | | Training |
| 33 | Saundatti | | Training |
| 34 | Sogi | | Training |
| 35 | Topaldoddi | | Training |
| 36 | Anadurwadi Tanda | | Training |
| 37 | Kaudiyal | | Training |
| 38 | Kajibilgi | | Testing |
| 39 | Ozhalapathy | Kerala | Training |
| 40 | Chadayangulay | | Training |
| 41 | Narasingam | | Training |
| 42 | Pasavadigomva | | Training |
| 43 | Pushpagiri | | Testing |
| 44 | Banbir Kheri | Madhya Pradesh | Training |
| 45 | Barkheri Bazar | | Training |
| 46 | Mamatkheda | | Training |
| 47 | Nachanbor | | Training |
| 48 | Nagda | | Training |
| 49 | Machaliya Ghat | | Training |
| 50 | Mandwa | | Training |
| 51 | Valiyarpani | | Testing |
| 52 | Ambral | Maharashtra | Training |
| 53 | Khokadi | | Training |
| 54 | Mahijalgaon | | Training |
| 55 | Raipur | | Training |
| 56 | Brahmanwel | | Training |
| 57 | Chakla | | Training |
| 58 | Jagmin | | Training |
| 59 | Kamravad | | Testing |
| 60 | Laimaton | Manipur | Training |
| 61 | Balesar | Rajasthan | Training |
| 62 | Kanod | | Training |
| 63 | Akal | | Training |
| 64 | Jhodal 2 | | Testing |
| 65 | Nayachar Island | West Bengal | Training |
| 66 | Nijkasba | | Testing |

(4) A comprehensive study on the optimum input parameters selection and data collection techniques is also needed to be examined for WS prediction accurately.

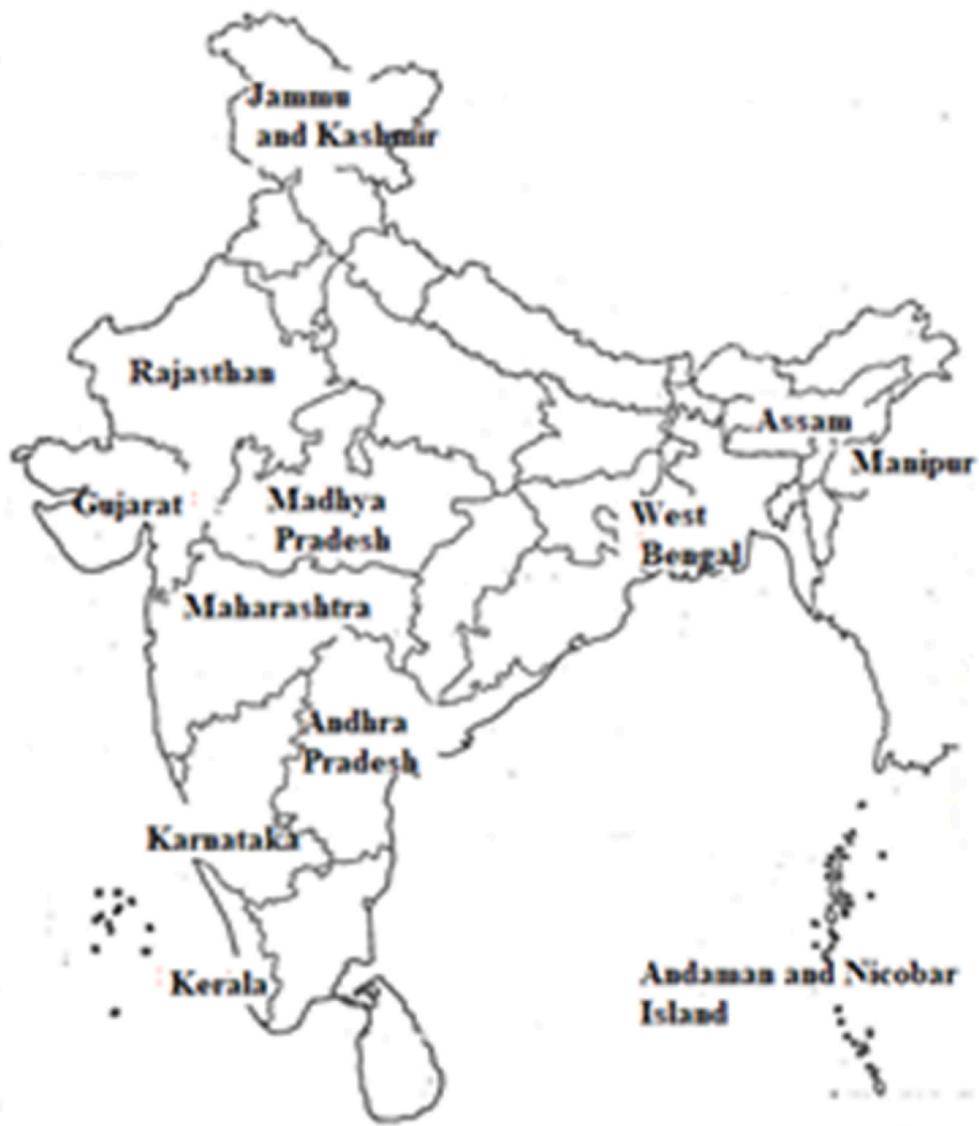

**Fig. 5.** Selected states of India for ANN model development.

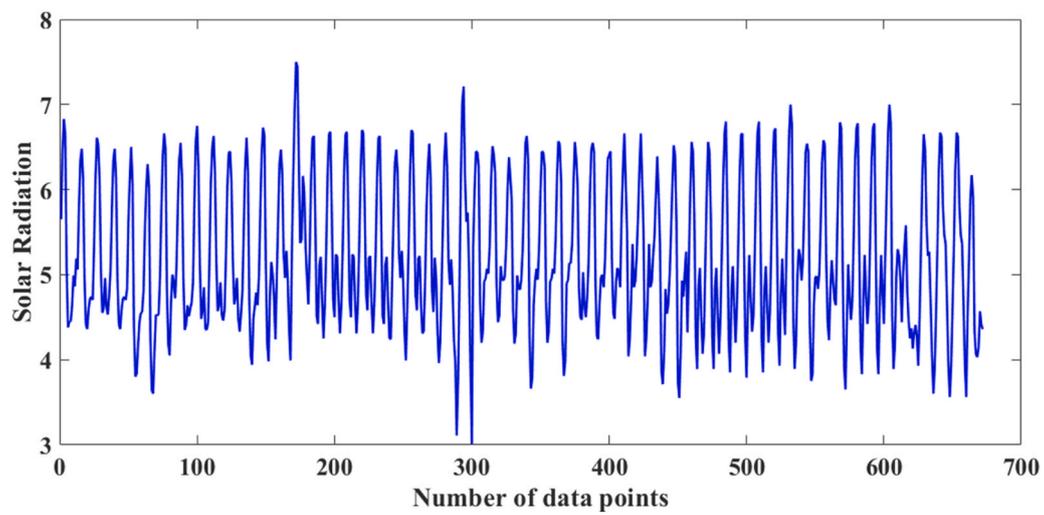

**Fig. 6.** Solar radiation (W/m$^2$).

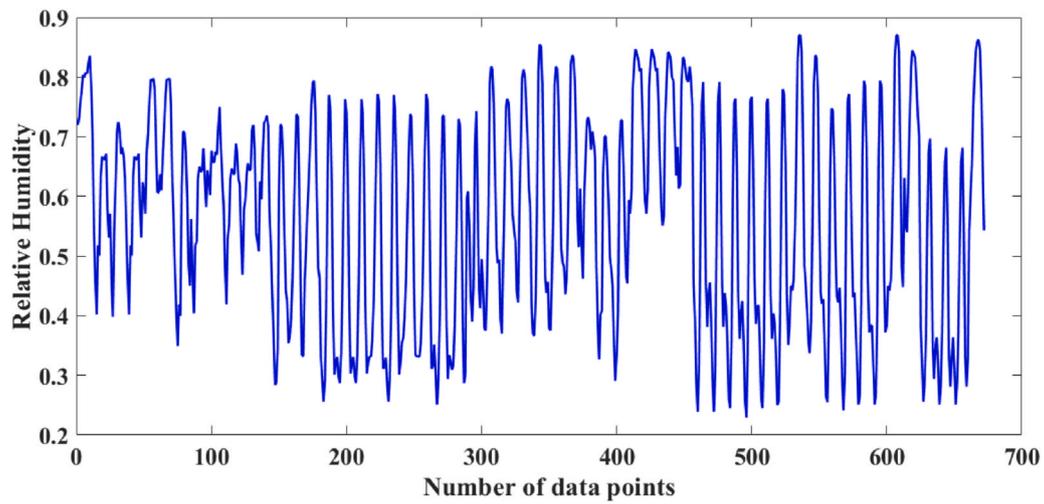

**Fig. 7.** Relative humidity.

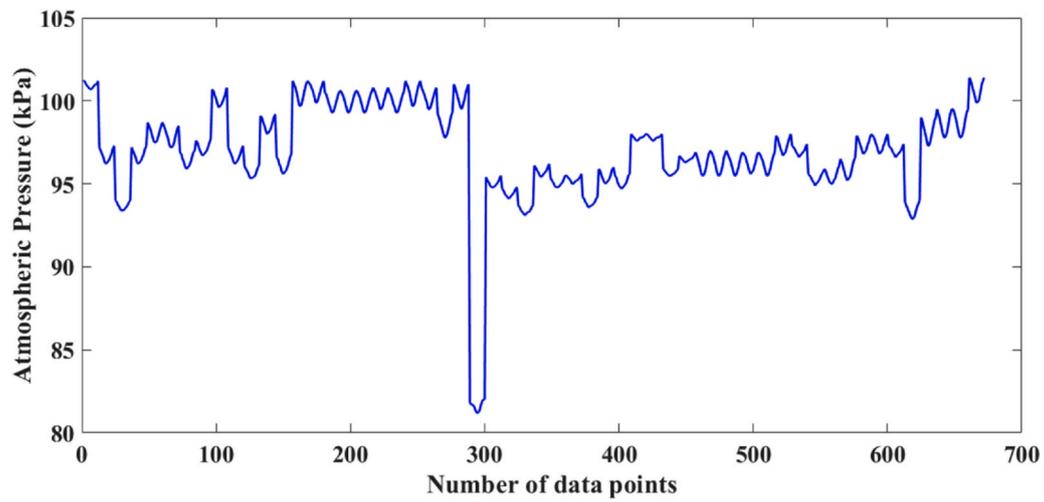

**Fig. 8.** Atmospheric pressure.

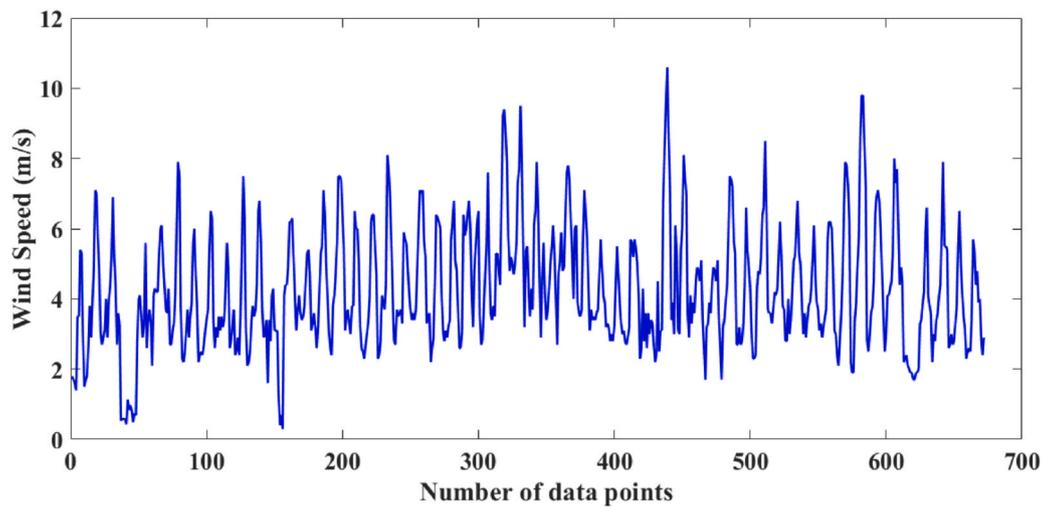

**Fig. 9.** Wind speed (m/s).

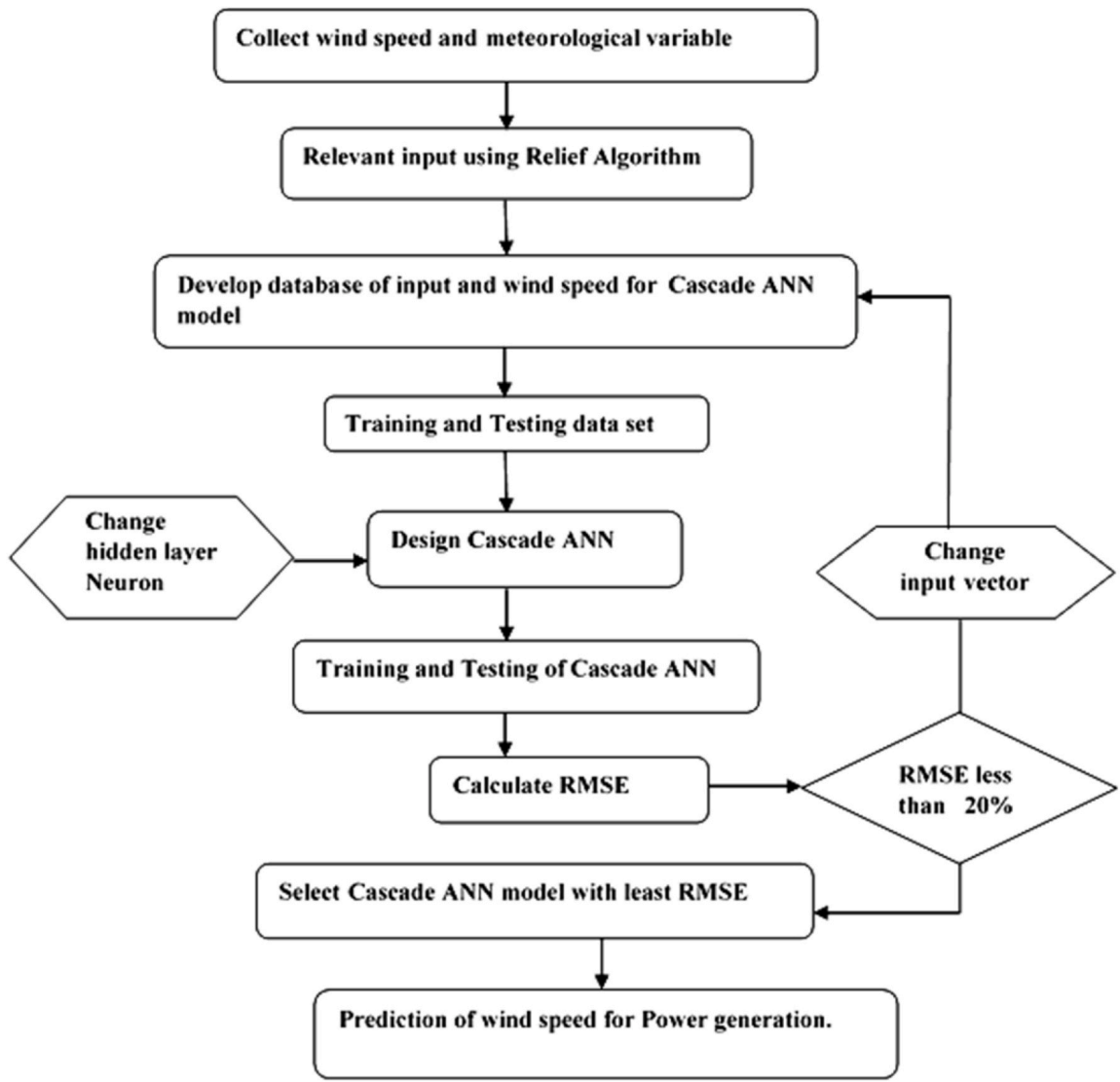

**Fig. 10.** Proposed algorithm for Wind Speed Prediction.

(5) Very few research has been done on finding the WS of a location where wind farm can be built i. e using data from other locations, wind speed of synoptic station is predicted.

(6) Further study on a combine approach of input data selection techniques and forecasting methods is needed.

(7) More hybrid techniques needs to be developed to increase forecasting accuracy.

(8) Relevant input variable selection using different feature selection techniques

## 5. Case study: feature selection techniques in wind speed prediction systems with the relief attribute algorithm

### 5.1. Description of locations

The Indian states which are considered to develop cascade ANN model are shown in Fig. 1 and data of different sites which are used for training and testing cascade ANN model is shown in Table 2. The 12 month average measured value of solar radiation, relative humidity, atmospheric pressure, temperature and wind speed for different sites shown in Figs. 5–9. These values are measured by National Aeronautics and Space Administration (NASA).

### 5.2. Relief Algorithm

Relief Algorithm (RA) is based on instant based learning and is given by Kira and Rendell [154] and later improved by Kononenko [155]. RA is very efficient in estimating attributes. It selects the features those are relevant to target. It locate nearest neighbour by random sampling a case from the data. The nearest neighbors attributes values are compared to sampled instance for updating relevance score for each attribute. For a given instances RELIEF searches two nearest neighbors: one from the identical class (nearest hit) and the other from different class (called nearest miss) [156]. RELIEF's estimate W [A] of attribute A is an approximation of the following difference of probabilities:

W [A] = P (different value of A\nearest instance from different class)-P (different value of A\nearest instance from same class).

The rationale is that good attribute should differentiate between instances from different classes and should have the same value for instances from the same class.

The algorithm developed in this study is shown in Fig. 10.

## 6. Results and discussions

To select relevant input variable Matlab code [rank] = relieff (j,i, 10) where j is a matrix of solar radiation, temperature, relative humidity, latitude, longitude, atmospheric pressure, earth temperature, elevation and i is measured wind speed. The ranks of input variables are shown in Table 3. It is found that atmospheric pressure, solar radiation, relative humidity and temperature are most relevant input variables and it can be used for prediction of wind speed.

**Table 3**
Rank of input variables.

| Input Variables | Rank |
| --- | --- |
| solar radiation | 2 |
| temperature | 4 |
| relative humidity | 3 |
| Latitude | 6 |
| longitude | 7 |
| atmospheric pressure | 1 |
| earth temperature | 5 |
| Elevation | 8 |

For training of ANN model 672 data points of atmospheric pressure, solar radiation and relative humidity are used as inputs and its corresponding monthly average wind speed are used as target. The sensitivity test is performed to validate the number of hidden layer neurons by calculating change in prediction error (MAPE) when number of hidden layer neurons is changed ±5 from hidden layer neurons. The performance plot of ANN model demonstrates that mean square error becomes minimum as number of epochs increases [Fig. 11]. The epoch is one complete sweep of training, testing and validation. The test set error and validation set error has comparable characteristics and no major over fitting happens near epoch 5 (where best validation performance has taken place). Epochs are one complete sweep of training, testing and validation. The developed Cascade ANN model is shown in Fig. 12. The error histogram plot [Fig. 13] shows error is minimum. The comparison between predicted and measured wind speed for training and testing cascade ANN model are shown in Fig. 14 and Fig. 15, showing predicted wind speed is close to measure wind speed. The root mean square error (RMSE) for training and testing are 1.44 m/s and 1.49 m/s respectively.

## 7. Conclusion

In this paper a review on different wind speed and wind power forecasting techniques is presented. Wind is an important energy source to meet our present and future energy requirement. Therefore, it is essential to forecast wind speed accurately. Various models have been developed and each model has its own characteristic. These models are based on different input variables, different locations, for different time horizons etc. so it is not possible to make accurate comparative analysis of these methods. Models based upon artificial intelligence methods predict wind speed with a low value of errors. Hybrid models outperforms historical methods of wind speed forecasting. Decomposition techniques can improve prediction accuracy of ANN models.

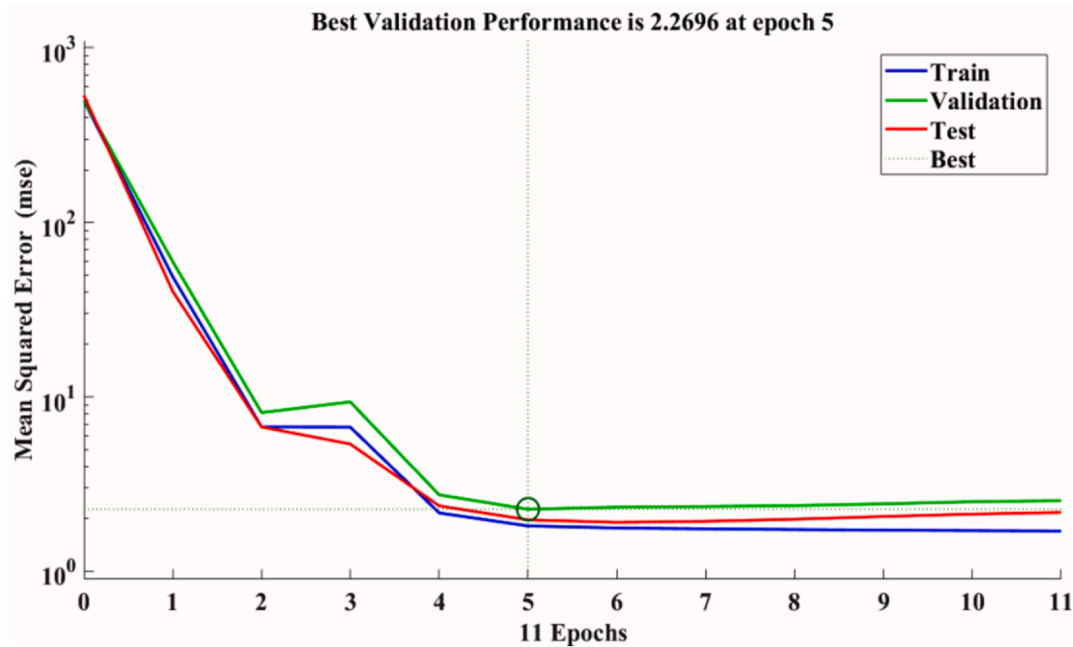

**Fig. 11.** Performance plot of Cascade ANN model.

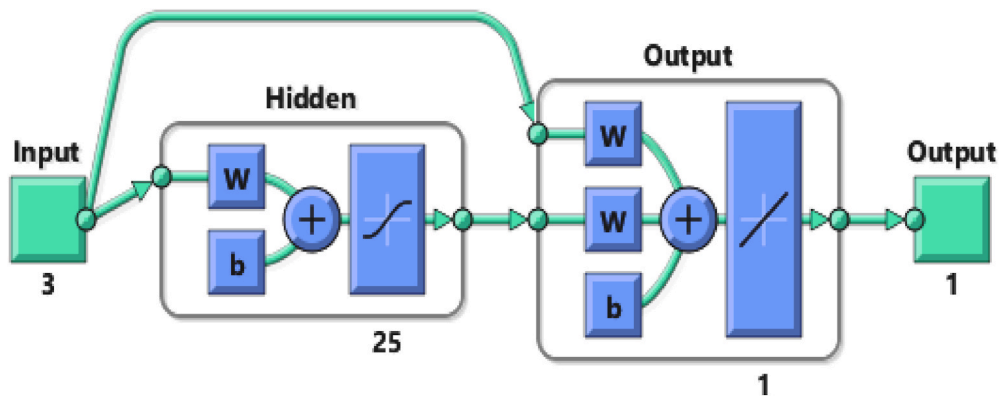

**Fig. 12.** Cascade ANN model.

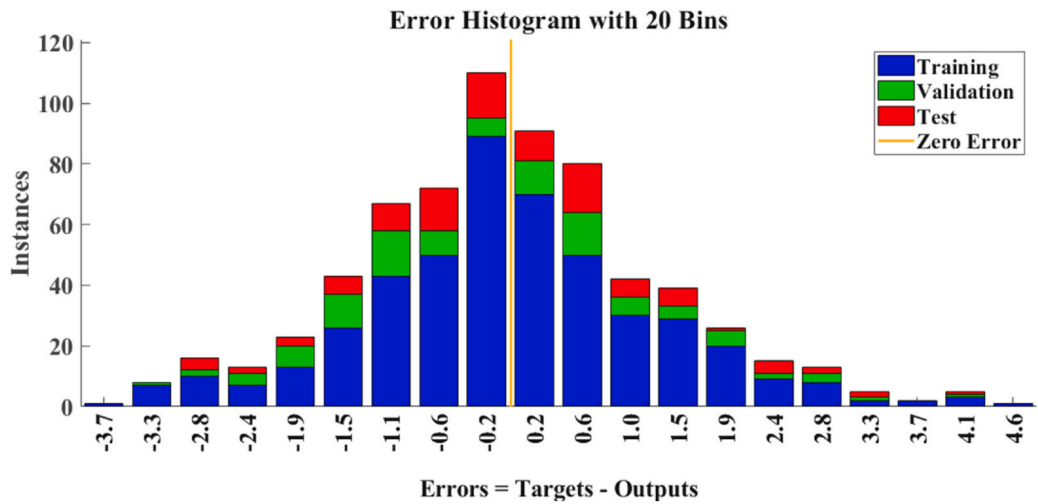

**Fig. 13.** Error histogram plot of Cascade ANN model.

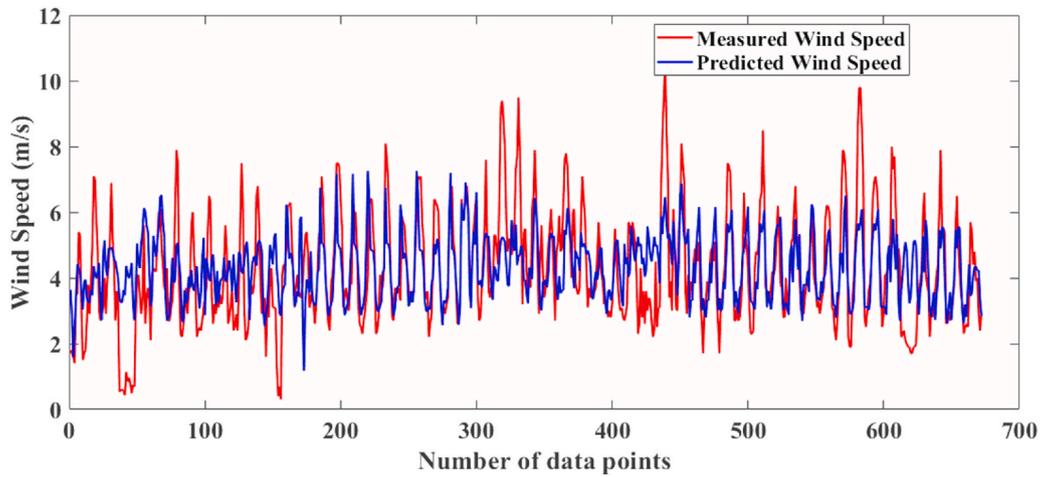

**Fig. 14.** Comparison between predicted and measured wind speed for training data points of Cascade ANN model.

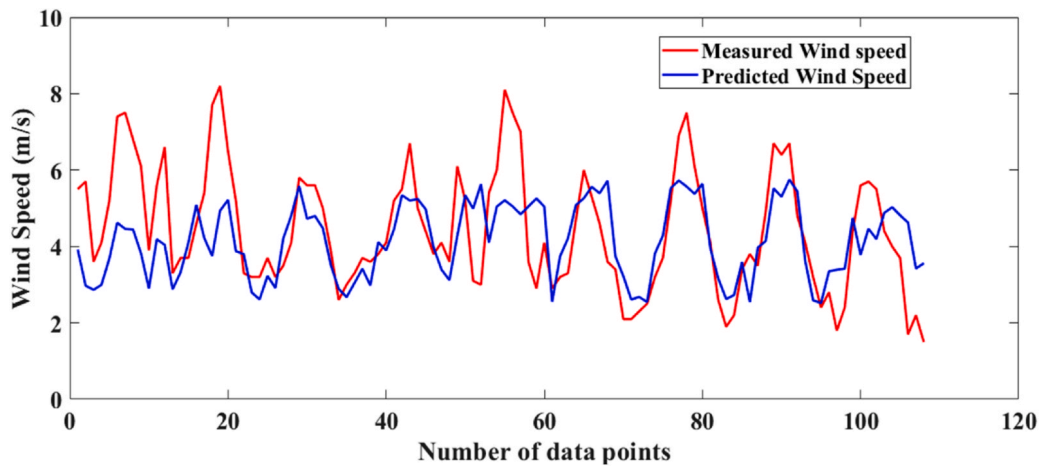

**Fig. 15.** Comparison between predicted and measured wind speed for testing data points of Cascade ANN model.

ANN with different algorithms and different number of neurons are discussed. Since training and testing of these ANN models is done with different data set values and different predictors so relative analysis of ANN methods is also not possible. It is important to consider wind speed, wind direction, temperature, pressure, relative humidity, vapor pressure and all other parameters that effect performance of prediction models. Also, small prediction error values are obtained if small variations in wind speed are considered and seasonal variations are eliminated. Therefore, relevant input parameters selections is prerequisite for accurate wind speed prediction.

As a case study based on research gap Relief algorithm using Matlab 2017 is implemented for identifying relevant input variables for the prediction of wind speed using Cascade ANN model. The most relevant input variables for the prediction of wind speed are found to be atmospheric pressure, solar radiation and relative humidity. The comparison between predicted and measured wind speed for training and testing data points, presents predicted wind speed is close to measure wind speed with root mean square error for training and testing are 1.44 m/s and 1.49 m/s respectively. The developed Cascade ANN model can be used to predict monthly average wind speed for other sites in India where wind speed measuring instruments are not installed, proving useful for assessment of wind power potential.

**Funding**

The work reported herewith has been financially by the Dirección General de Universidades, Investigación e Innovación of Castilla-La Mancha, under Research Grant ProSeaWind project (Ref.: SBPLY/19/180501/000102) and the ERDF funds.

**Declaration of competing interest**

The authors declare that they have no known competing financial interests or personal relationships that could have appeared to influence the work reported in this paper.

**Acknowledgement**

The authors would like to acknowledge the support from Intelligent Prognostic Private Limited India Researcher's Supporting Project. The authors would like to acknowledge the support from Ingenium Research Group, Universidad Castilla-La Mancha, 13071 Ciudad Real, Spain. The authors would like to acknowledge the support from Department of Quantitative Methods, CUNEF Universidad, 28040 Madrid, Spain. The authors would like to acknowledge National Aeronautics and Space Administration for providing measured value of wind speed and other meteorological variables open access for the study.